# Intelligent Personal Assistant with Knowledge Navigation

Amit Kumar, Computer Science Student at AIT Pune
Rahul Dutta, Computer Science Student at AIT Pune
Harbhajan Rai, Computer Science Student at AIT Pune
*Under Guidance of Prof. Rushali Patil, Computer Science Department, AIT Pune*

**Abstract**

An Intelligent Personal Agent (IPA) is an agent that has the purpose of helping the user to gain information through reliable resources with the help of knowledge navigation techniques and saving time to search the best content. The agent is also responsible for responding to the chat-based queries with the help of Conversation Corpus. We will be testing different methods for optimal query generation. To felicitate the ease of usage of the application, the agent will be able to accept the input through Text (Keyboard), Voice (Speech Recognition) and Server (Facebook) and output responses using the same method. Existing chat bots reply by making changes in the input, but we will give responses based on multiple SRT files. The model will learn using the human dialogs dataset and will be able respond human-like. Responses to queries about famous things (places, people, and words) can be provided using web scraping which will enable the bot to have knowledge navigation features. The agent will even learn from its past experiences supporting semi-supervised learning.

1. Introduction

    Whenever we have a conversation, each response is based on the previous sentence heard. If we have any dataset of human conversations then the same rule applies to that dataset. This theory is true for more than 95% of the conversations. The basic idea of developing the bot which can respond on facebook was to enable as many users as possible to interact with the bot making a large learning database for future references.

    In this paper we try to find the best possible matching line in the database to the input query. We will check the different algorithms for analyzing the dataset to enable faster search operations. MongoDB will be used as the backend database as it provides faster read operations. The database will update itself after each conversation and use this knowledge in the next chat.

2. Previous Works

    Existing chatter bot systems have a few issues which should be removed in order to make the best model. Initial bots used to respond by making changes in the input (like replacing "I" with "You" and "We" with "They" and vice-versa). Example:

    *Input: I want to know this.     Output: You want to know this.*

    This type of response by the bot isn't the best and a better response is provided below.

    *Input: I want to know this.     Output: What do you what to know?*

    Another instances where these bots fail is the case of answering the common quiz based questions like:

    *Input: Who is Sachin Tendulkar?    or    Input: What is DHCP?    or    Input: When is Independence Day celebrated?*

We will add the features of web scraping which will enable the bot to answer these queries and behave more user friendly. Many other bots even tend to ask questions from the users instead of replying to their queries which tend to make the Turing Test impenetrable for these bots.

3. **The Learning Phase**

   *3.1 The Corpus of Conversations*

   We have used a corpus of Friends Series Dataset with SRT files of 184 episodes consisting of more than 75000 lines of human conversations. Each line in the conversation corpus is related to the next line of the Dataset. We plan to find the most matching line in the dataset with the input query and reply with the next line of the dataset.

   *3.2 Pre-processing of the Corpus*

   The corpus was studied and vector for each line was created which will be used using Similarity Check.
   The noisy lines consisting of blanks, starting and ending time of the lines and narrators lines are removed so that the bot can learn from noise-free dataset of conversations.

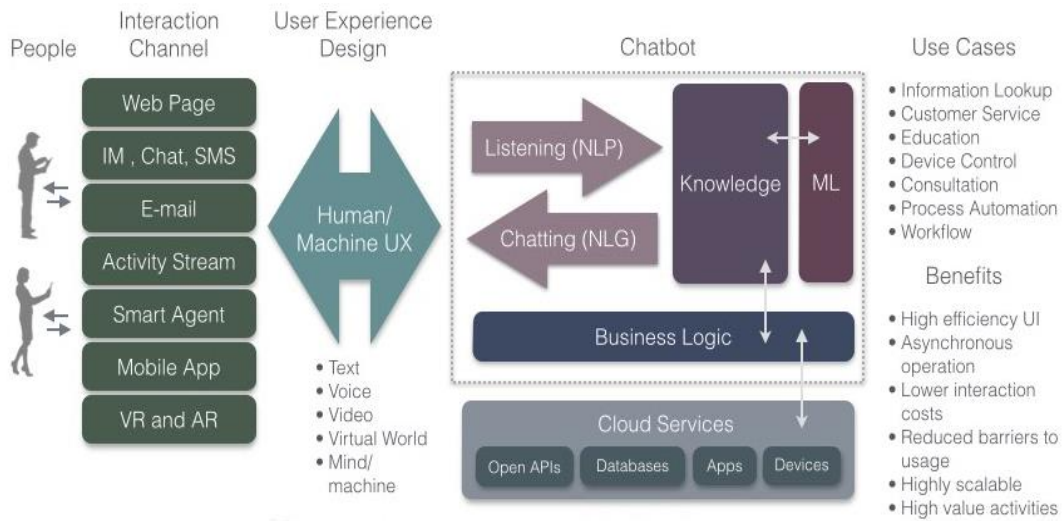

Fig 1. General Architecture

4. **Algorithm Selection**

   For searching a particular sentence in the database, we can use the most number of words matching in the input and the output query. We will use NLTK for removing the infected forms of the same word. Like "pick", "picked", "picks" are all different inflected forms of the word "pick". This process (lemmatization) can be performed using WordNet corpora from NLTK in Python very easily. We will remove the instances of words with length lesser than three (like "a", "an", "on"). We will even remove the words pertaining to the meaning and grammar of the sentence. These are words like "above", "are", "the", "beneath" and "around". We can then calculate the distance between the input query and each possible output query. The query which is at the least distance from the input query will be considered to generate the response.

   The issue with this kind of approach is that it neglects any kind of grammar or any meaning associated and the result might be very different from what was originally expected.
   Example:

*Input: I am gathering things for the picnic.     Input: There were many people at the gathering.*

In the first sentence *gathering* is a verb and its original replacement should be *gather* while in the second case *gathering* is a noun and its replacement should be *gathering* itself.

Instead of keeping the track of which words appear, we will also have count of number of occurrences of the each word which will make the distance calculation much more reliable. To calculate the distance between these strings we can use L1 or L2 normalized forms.

Another approach will be calculating the Levenstein Distance which makes use of Dynamic Programming in order to calculate the number of changes required in the input string to convert it into the output string. While the complexity of this algorithm tend to be O(n*j*k), it still works faster than O(n) time complexity. Where, j and k are the lengths of the input and the output string respectively. We use the fact that any input string will have length not more than 30-40. Thus k*l will be constant factor in the operation and will not add any overhead cost in the complexity. This way we convert the $O(n^3)$ complexity approach into O(n) time complexity. Moreover we can introduce parallel programming to search in the database so that multiple lines can be searched at once.

In order to answer the knowledge based questions, we make use of the fact that similar types of queries will have similar form of response from the web. We handle one input for each type of query and other similar queries can itself be handled by the bot.

5. **Conclusion**

We find that the Levenstein based approach can handle the tasks optimally. This even takes care of the ordering of the words in the sentence and even the grammar associated with it. Using parallelism will even reduce the overhead cost by a large margin. We will compare the learning time using Levenstein and the Similarity Based method. A faster approach can be thought of using hashing which will use the Similarity Based approach by saving the frequencies of each word in a vector and then using unordered map or set for constant time searching.